\documentclass{article}
\PassOptionsToPackage{authoryear}{natbib}
\usepackage{amsmath}
\usepackage{graphicx}
\usepackage{algorithm}
\usepackage{algpseudocode}
\usepackage[preprint]{neurips_2025}
\usepackage[utf8]{inputenc} 
\usepackage[T1]{fontenc}    
\usepackage{hyperref}       
\usepackage{url}            
\usepackage{booktabs}       
\usepackage{amsfonts}       
\usepackage{nicefrac}       
\usepackage{microtype}      
\usepackage{xcolor}         
\title{Deterministic Continuous Replacement: Fast and Stable Module Replacement in Pretrained Transformers}
\author{
  Rowan Bradbury \\
  Bradbury Group \\
  \texttt{rowan@bradburygroup.org} \\
  \And
  Aniket Srinivasan Ashok \\
  Bradbury Group \\
  University of Waterloo \\
  \texttt{aniket@bradburygroup.org} \\
  \And
  Sai Ram Kasanagottu \\
  Bradbury Group \\
  SUNY Stony Brook \\
  \texttt{sairam@bradburygroup.org} \\
  \And
  Gunmay Jhingran \\
  Bradbury Group \\
  Delhi Technological University \\
  \texttt{gunmay@bradburygroup.org} \\
  \And
  Shuai Meng \\
  Bradbury Group \\
  UC Berkeley \\
  \texttt{shuai@bradburygroup.org} \\
}
\begin{document}
\maketitle
\begin{center}
    \vspace{-0.6em}
    \textit{Accepted to the NeurIPS 2025 ScaleOPT Workshop.}
\end{center}
\vspace{0.8em}

\begin{abstract}
  Replacing modules in pretrained models—especially swapping quadratic self-attention for efficient attention alternatives—poses a hard optimization problem: cold-start reinitialization destabilizes frozen backbones. We isolate this core stability challenge in a controlled study. \textbf{Deterministic Continuous Replacement (DCR)} blends teacher and student outputs with a deterministic, annealed weight $\alpha(t)$. Theoretically, DCR eliminates gate-induced gradient variance inherent to stochastic replacement (Sec.~\ref{sec:theory}). In a single-seed study, DCR attains faster convergence and stronger alignment than stochastic gating and distillation baselines on controlled attention replacement, establishing a foundation for heterogeneous operator swaps.
\end{abstract}
\section{Introduction}
As training costs rise, model adaptation has become critical. Two trends converge: compression pipelines replace blocks with smaller surrogates \citep{han2015deep, sanh2019distilbert}, and efficient attention variants \citep{wang2020linformer, choromanski2020rethinking, beltagy2020longformer} promise $O(n)$ or $O(n \log n)$ complexity. However, replacing modules with cold-start operators inside frozen backbones destabilizes optimization: downstream blocks receive out-of-distribution features, leading to optimization instability, ineffective gradient updates, and slow recovery. 
Existing approaches face fundamental tradeoffs. Knowledge distillation methods \citep{hinton2015distilling, romero2014fitnets} require expensive teacher forward passes and enforce rigid feature matching, while stochastic replacement strategies like BERT-of-Theseus \citep{xu2020bert} introduce gradient variance and uneven recovery. We isolate the \emph{core replacement stability problem}: integrating a randomly initialized module into a frozen backbone—the optimization challenge common to all replacement scenarios. By studying this in a controlled setting (replacing attention with re-initialized attention), we eliminate representational mismatch as a confound, allowing us to attribute differences solely to stability mechanisms and rigorously measure gradient dynamics.

\textbf{Our contributions.}
\begin{itemize}
    \item \textbf{DCR method} (Sec.~\ref{sec:method}): deterministic blending that eliminates gate-induced gradient variance and naturally enables near-zero-cost feature alignment since both teacher and student outputs are computed for the blend.
    \item \textbf{Variance reduction theory} (Sec.~\ref{sec:theory}): formal analysis isolating and eliminating the gate-induced variance term inherent to stochastic replacement (Props.~1--2), with bounds on curvature bias and loss-path smoothness.
    \item \textbf{Controlled validation} (Sec.~\ref{sec:experiments}): faster convergence and stronger alignment than stochastic gating and distillation baselines in a controlled self-replacement setting that isolates stability from representational mismatch.
\end{itemize}

While experiments focus on controlled self-replacement on smaller models (CIFAR-100, ViT-Small) to enable rigorous ablation, the method and theory are explicitly constructed for heterogeneous operator swaps (e.g., Linformer, Performer, sparse/Fourier attention), which is the immediate follow-on work. We present this study in workshop format because isolating and formalizing the replacement-stability gap is a prerequisite for scalable deployment in production settings, where understanding failure modes and convergence guarantees is critical. DCR's efficiency advantage over distillation is amplified in compute-saturated regimes—large language model or diffusion transformer replacement—where GPU utilization is high and the full teacher model forward pass required by distillation directly increases wall-clock cost relative to DCR's branch-local teacher evaluation.

\section{Related Work}

\paragraph{Knowledge Distillation and Model Compression.}
Knowledge distillation \citep{hinton2015distilling} and feature-based variants \citep{romero2014fitnets, touvron2021training} require a full separate teacher model forward pass per training step and impose rigid alignment constraints. In contrast, DCR computes teacher outputs only at the replaced modules, avoiding full-model duplication. Compression methods \citep{han2015deep, sanh2019distilbert, jiao2019tinybert} assume parameter compatibility, which breaks down for heterogeneous operators.

\paragraph{Stochastic Module Replacement.}
BERT-of-Theseus \citep{xu2020bert} randomly selects between teacher and student modules via a Bernoulli gate $z_\ell(t) \sim \mathrm{Bernoulli}(p(t))$, enabling gradual knowledge transfer without explicit feature matching. However, this introduces gradient variance in the student-only gradient:
\[
\nabla_{\theta_\ell} L = z_\ell \cdot J_{G_\ell}^\top \frac{\partial S_\ell}{\partial \theta_\ell},
\]
producing high variance when $p(t)$ is mid-range. We replace this stochastic gate with a deterministic blend. For controlled ablation, we introduce \textbf{Theseus-Gumbel}, using soft Gumbel-Softmax gates $r_\ell(t) \in (0,1)$ with temperature $\tau$:
\[
r_\ell(t) = \mathrm{GumbelSoftmax}(p(t), \tau),
\]
\[
x_{\ell+1} = x_\ell + r_\ell(t) S_\ell(h_\ell) + (1 - r_\ell(t)) T_\ell(h_\ell).
\]
This allows gradients to flow but retains gate-induced variance (GUM, GUM+DFG baselines in experiments).

Table~\ref{tab:method-comparison} summarizes the key differences between DCR and prior module replacement approaches across three critical dimensions: gradient variance, computational overhead, and feature matching requirements.

\begin{table}[h]
\centering
\caption{Comparison of module replacement methods. DCR achieves low gradient variance and minimal branch-local overhead \textbf{in our non-compute-saturated experimental regime}, addressing the stability bottleneck that limits both distillation and stochastic replacement.}
\label{tab:method-comparison}
\resizebox{\textwidth}{!}{%
\begin{tabular}{lccc}
\toprule
Method & Gradient Variance & Extra Compute & Needs Feature Matching \\
\midrule
Knowledge Distillation & Low & High (full teacher model forward) & Yes (rigid) \\
Theseus (Stochastic) & High (gate-induced) & Low (gate overhead only) & No \\
DCR (Ours) & Low (deterministic) & Low (branch-local teacher only at replaced layers) & No (optional via DFG) \\
\bottomrule
\end{tabular}%
}
\end{table}

\paragraph{Replacement Stability Gap.}
Prior methods assume parameter continuity; under cold-start reinitialization, downstream layers receive out-of-distribution features. \textbf{DCR targets this gap} with a deterministic, low-variance path stable under reinitialization.

\section{Methodology}
\label{sec:method}
\subsection{Problem Formulation}
Given pretrained network $F$ with $L$ modules, we replace subset $\mathcal{I}\subseteq\{1,\dots,L\}$. For $\ell\in\mathcal{I}$, let $T_\ell$ (frozen teacher) and $S_\ell(\cdot;\theta_\ell)$ (trainable student) share input/output shapes. Denote normalized input $h_\ell=\mathrm{LN}(x_\ell)$ and frozen tail $G_\ell$. DCR blends on the residual branch:
\begin{equation}
\label{eq:dcr-path}
x_{\ell+1}(t) \;=\; x_\ell(t) \;+\; \big[\alpha(t)\,T_\ell(h_\ell(t)) \;+\; (1-\alpha(t))\,S_\ell(h_\ell(t);\theta_\ell)\big],
\end{equation}
with global gate $\alpha(t)\in[0,1]$: $\alpha(0)=1$ (teacher-only) $\to$ $\alpha(T)=0$ (student takeover).
\subsection{Theoretical Properties (Stability)}
\label{sec:theory}
\paragraph{Analysis scope.}
We analyze gate-induced variance and path geometry under frozen $G_\ell$ and scheduled gates, holding the student's input distribution fixed at each step. This provides local, conditional intuition—not full training-dynamics guarantees or global convergence proofs—validated empirically in Sec.~4. Standard assumptions: teacher runs in \texttt{eval()} mode, gates independent of minibatch, differentiable functions.

\paragraph{Lower gate-induced gradient variance.}
To our knowledge, this is the first formulation that analytically isolates and eliminates the gate-induced variance term central to stochastic replacement. Let
\[
a \;:=\; J_{G_\ell}^\top \frac{\partial S_\ell}{\partial \theta_\ell}\ \in \mathbb{R}^{\dim(\theta_\ell)}.
\]

\noindent\textbf{Proposition 1 (Variance decomposition for Theseus).}
If a hard gate $z\sim\mathrm{Bernoulli}(p)$ selects student vs.\ teacher (independent of the data), then
\[
\nabla_{\theta_\ell} L \;=\; z\,a,\qquad
\mathbb{E}[\nabla_{\theta_\ell}L] = p\,\mathbb{E}[a],\qquad
\]
\[
\mathrm{Var}[\nabla_{\theta_\ell}L] \;=\; p\,\mathrm{Var}[a] \;+\; p(1-p)\,\|\mathbb{E}[a]\|^2
\;\le\; p\,\mathrm{Var}[a] + p(1-p)\,\mathbb{E}\|a\|^2.
\]
This decomposition reveals the gate-induced variance component that DCR eliminates. \emph{(Proof in Appendix~\ref{app:proof-prop1}.)}

\medskip
\noindent\textbf{Proposition 2 (Deterministic gate removes gate-induced variance).}
Let $Y_\alpha := (1-\alpha)\,T_\ell + \alpha\,S_\ell$ and define $a(y) := J_{G_\ell}(y)^\top \frac{\partial S_\ell}{\partial \theta_\ell}$, so that the DCR gradient is
\[
\nabla_{\theta_\ell} L_{\mathrm{DCR}} \;=\; \alpha\,a(Y_\alpha),
\]
whereas under Theseus (hard gate $z\!\sim\!\mathrm{Bernoulli}(p)$) the student gradient is $\,\nabla_{\theta_\ell} L_{\mathrm{Th}} = z\,a(S_\ell)$. Let $X$ denote the minibatch. Then the \emph{gate-induced} component of gradient variance,
\[
\mathbb{E}\!\left[\mathrm{Var}\!\big(\nabla_{\theta_\ell} L \,\big|\, X\big)\right],
\]
is zero for DCR and equals $p(1-p)\,\mathbb{E}\|a(S_\ell;X)\|^2$ for Theseus. Hence,
\[
\mathbb{E}\!\left[\mathrm{Var}\!\big(\nabla_{\theta_\ell} L_{\mathrm{Th}} \,\big|\, X\big)\right]
\;-\;
\mathbb{E}\!\left[\mathrm{Var}\!\big(\nabla_{\theta_\ell} L_{\mathrm{DCR}} \,\big|\, X\big)\right]
\;=\; p(1-p)\,\mathbb{E}\|a(S_\ell;X)\|^2 \;\ge\; 0.
\]
This is the core theoretical justification for DCR: strictly lower gradient variance than stochastic gating. \emph{(Proof in Appendix~\ref{app:proof-prop2}.)}

\medskip
\noindent\textbf{Remark (Soft gates / Theseus-Gumbel).}
Let $r\in(0,1)$ be a random soft gate with $\mathbb{E}[r]=p$, $\mathrm{Var}(r)>0$ (e.g., Gumbel-Softmax, temperature $\tau$). Then
\[
\mathrm{Var}[r a]
= \mathbb{E}[r^2]\mathrm{Var}[a] + \mathrm{Var}(r)\,\|\mathbb{E}[a]\|^2
= p^2\,\mathrm{Var}[a] \;+\; \underbrace{\mathrm{Var}(r)\,\mathbb{E}\|a\|^2}_{\text{extra, gate-induced}},
\]
so any stochastic gate incurs an additional nonnegative term that DCR does not.

\paragraph{Curvature bias.}
Stochastic mixing through nonlinearities introduces a curvature-dependent bias: the expected output after a nonlinearity differs from the nonlinearity applied to the expected blend. DCR's deterministic path avoids this entirely (Proposition~3, Appendix~\ref{app:prop3}).

\paragraph{Summary.}
Props.~1--2 show DCR removes gate-induced variance. Proposition~3 (Appendix~\ref{app:prop3}) shows DCR avoids curvature bias from stochastic mixing through nonlinearities. Proposition~4 (Appendix~\ref{app:prop4}) bounds the loss path via Lipschitz continuity. Under the stated assumptions (independent gate scheduling, fixed frozen tail, local smoothness), deterministic blending yields a better-conditioned optimization path. These results are local and conditional, not full training-dynamics guarantees. Empirical validation in Sec.~\ref{sec:experiments}. Algorithm~\ref{alg:dcr_process} (Appendix) details the full procedure.
\subsection{Deep Feature Guidance (DFG)}
\label{sec:dfg}
While DCR ensures a smooth replacement, the student can benefit from direct interface alignment. Since DCR already evaluates both $T_\ell(h_\ell)$ and $S_\ell(h_\ell)$ for the blend at replaced layers, the auxiliary loss adds near-zero marginal cost—no additional forward passes are required. This contrasts with standard knowledge distillation, which requires a full teacher model pass. We add an auxiliary loss on the residual outputs at the replaced sites:
\begin{equation}
\label{eq:dfg-loss}
\mathcal{L}_{\mathrm{DFG}}
\;=\;
\sum_{\ell \in \mathcal{I}} \big\| S_\ell(h_\ell) - T_\ell(h_\ell) \big\|_2^2,
\qquad
h_\ell = \mathrm{LN}(x_\ell).
\end{equation}
Let $\hat{y}=F_t(x)$ denote the model output under the current global gate $\alpha(t)$. The overall objective is
\begin{equation}
\label{eq:total-loss}
\mathcal{L}_{\mathrm{total}}
\;=\;
\mathcal{L}_{\mathrm{task}}(\hat{y},\,y^\star) \;+\; \lambda\,\mathcal{L}_{\mathrm{DFG}},
\qquad \lambda \ge 0
\end{equation}
Since DCR already computes both $T_\ell(h_\ell)$ and $S_\ell(h_\ell)$ for blending, DFG adds negligible cost ($\lambda \ge 0$ controls strength). In our non-compute-saturated regime, DCR's overhead scales with the number of replaced modules $|\mathcal{I}|$, whereas knowledge distillation incurs a full teacher forward regardless of $|\mathcal{I}|$. We anneal $\lambda$ following the same aggr20 schedule as $\alpha$; full schedule in Appendix~\ref{ss:appendixexp}. For Theseus-Gumbel+DFG, we evaluate the teacher branch locally even when the gate selects the student.
\section{Controlled Evaluation of Replacement Stability}
\label{sec:experiments}
\subsection{Experimental Setup}
\paragraph{Datasets and Models.}
We evaluate DCR on ImageNet-pretrained ViT Small models \citep{dosovitskiy2020image}, serving as the teacher backbone finetuned on CIFAR100 \citep{krizhevsky2009learning}. Student modules replace attention blocks and are randomly re-initialized (Kaiming initialization \citep{DBLP:journals/corr/HeZR015}). All DCR blending is applied post-softmax and prior to residual addition.

\paragraph{Replacement Schedules.}
DCR (aggr20): $\alpha$ transitions $1.0\to0.0$ over first 20\% of training. Theseus variants use inverse probability $p$. DFG anneals with aggr20 schedule.
\subsection{Results}
\begin{figure}[h]
    \centering
    \includegraphics[width=\columnwidth]{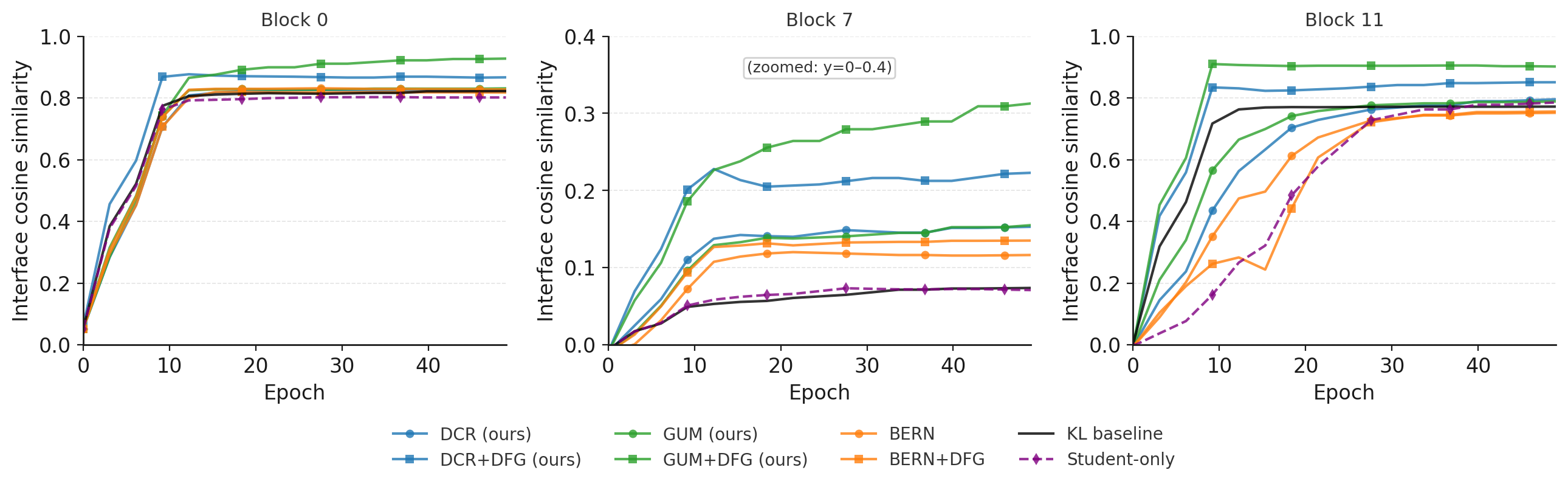}
    \caption{Interface cosine similarity (cosine similarity of residual outputs) between teacher and student outputs at different layers (Block 0, 7, 11) across training epochs.}
    \label{fig:dcr_infer}
\end{figure}

\paragraph{Alignment Dynamics.}
DCR and DCR+DFG achieve consistently higher interface cosine similarity than stochastic baselines (Figure~\ref{fig:dcr_infer}), with largest gains in mid and late blocks. Crucially, deterministic blending ensures downstream blocks receive in-distribution features from the start, enabling later layers to learn earlier without wasted gradients—avoiding the plateauing seen in GUM and BERN where gate-induced starvation delays deep-layer convergence.

\begin{figure}[h]
    \centering
    \includegraphics[width=\columnwidth]{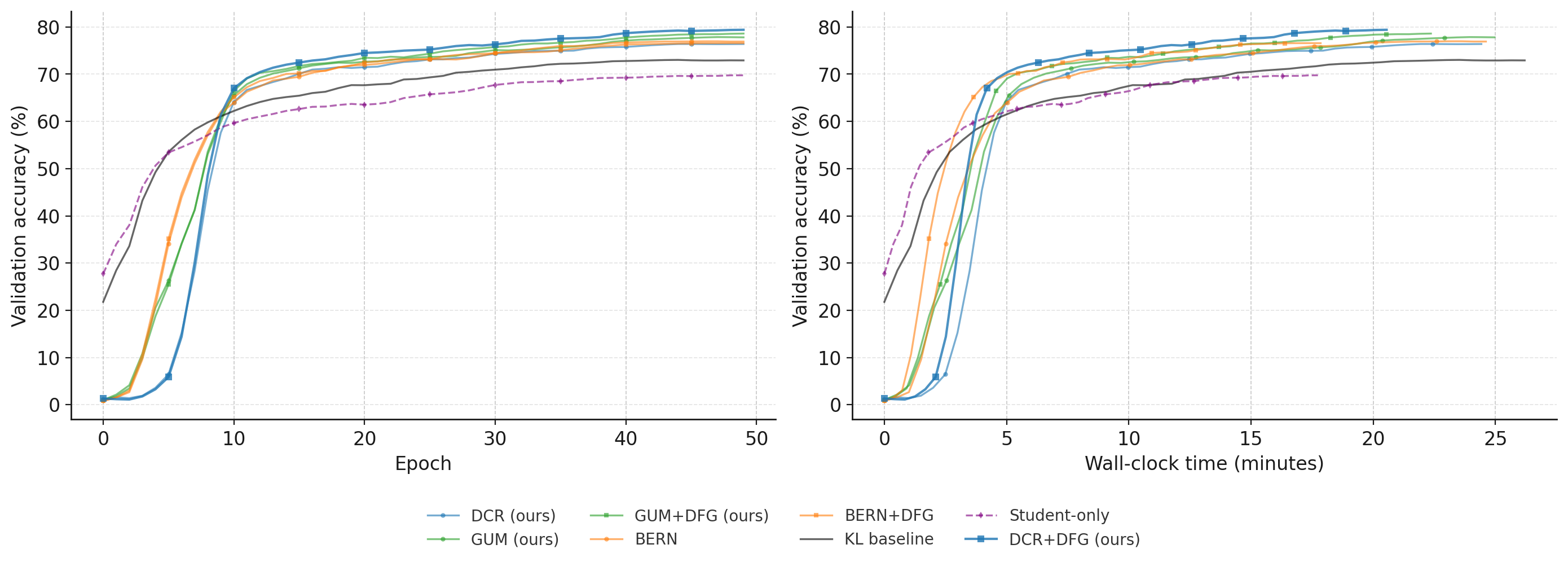}
    \caption{Validation accuracy during module replacement on CIFAR-100 (ViT-Small/16). Left: epochs. Right: wall-clock time.}
    \label{fig:valid_acc}
\end{figure}

\paragraph{Accuracy Recovery and DFG Effect.}
DCR variants reach target accuracy sooner in both epoch and wall-clock views (Figure~\ref{fig:valid_acc}), despite similar final accuracies ($\approx$78--80\%). Adding DFG accelerates takeover without full teacher passes, with strongest gains in deeper blocks—confirming near-zero-cost feature guidance compounds with deterministic blending.
\section{Research Positioning and Scope}
\label{sec:limitations}
Our controlled design prioritizes internal validity: single model family (ViT-Small), dataset (CIFAR100), single seed, I/O-bound regime, pre-norm residual Transformers, and self-replacement (attention $\to$ re-initialized attention) to isolate stability from representational mismatch. Intentional scoping choices include: (i) global gate $\alpha(t)$ rather than per-layer or progressive schedules, (ii) comparison to Theseus variants and student-only baselines without function-preserving initialization (Net2Net) or stronger alignment methods (CKA, Gram matching, learned adapters), (iii) no exhaustive hyperparameter tuning. Architectures with batch normalization, extensive simultaneous replacements across many layers, or other normalization schemes may exhibit different stability dynamics. These constraints enable causal attribution of variance reduction effects—rarely possible when varying architecture, operators, and compute simultaneously—and establish methodological clarity as a prerequisite for scaling to heterogeneous operators. Results should be interpreted as feasibility evidence rather than definitive benchmarking. Extensions to compute-saturated regimes, heterogeneous operators, and diverse architectures are natural next steps.
\section{Conclusion}
We introduced \textbf{Deterministic Continuous Replacement (DCR)}, which eliminates gate-induced gradient variance in cold-start module replacement. In controlled experiments, DCR+DFG outperforms stochastic gating and distillation baselines, establishing a foundation for heterogeneous operator swaps. For frozen-backbone replacement under our assumptions, DCR provides a stable, efficient alternative to stochastic methods. Next steps include: heterogeneous operators (efficient attention variants), larger models, compute-saturated regimes, and per-layer adaptive $\alpha$ schedules conditioned on interface similarity for deep architectures.

\paragraph{Code Availability.}
Code will be released in a future extended version of this work.

\section*{Acknowledgements}
This work was conducted by the Bradbury Group, an independent non-profit AI research lab. Beyond the listed authors, we thank the broader Bradbury Group research community for internal reviews, discussion of early prototypes, and support with maintaining the shared training and evaluation infrastructure, with special thanks to Elea Zhong and Joseph Haynes for their detailed manuscript review and feedback. This project received no external funding and was enabled entirely through the lab's internal resources.

\bibliographystyle{unsrtnat} 
\bibliography{references} 
\newpage
\appendix
\section{Technical Appendices and Supplementary Material}
\subsection{Detailed Experimental Setup}\label{ss:appendixexp}
\paragraph{Training Procedure.}  
Experiments are conducted on NVIDIA A100 GPUs with mixed-precision (BF16) on CIFAR100 dataset. Training follows two stages: (i) \textbf{Head warmup}: only the classification head is trained for 6 epochs with 2 epoch warmup and lr $1\times10^{-3}$, cosine annealing to $1\times10^{-6}$ full weights unfrozen and further finetune at $1\times10^{-4}$ for 6 further epochs, label smoothing 0.1, and mixed precision; (ii) \textbf{Full model training}: all layers unfrozen, base lr $5\times10^{-4}$ with cosine annealing over 50 epochs, weight decay 0.05, gradient clipping 1.0, batch size 128, AdamW optimizer \citep{DBLP:journals/corr/abs-1711-05101}(eps $1\times10^{-8}$, betas $[0.9,0.999]$), and label smoothing 0.1.
\paragraph{Replacement Schedules.}  
For DCR (aggr20), $\alpha$ transitions from 1.0 to 0.0 over the first 20\% of training: Phase 1 (0--10\%) $\alpha=1.0\to0.3$; Phase 2 (10--20\%) $\alpha=0.3\to0.0$; Phase 3 (20--100\%) $\alpha=0.0$. Stochastic Theseus variants follow the inverse probability $p$: Phase 1 (0--10\%) $p=0.1\to0.7$; Phase 2 (10--20\%) $p=0.7\to1.0$; Phase 3 (20--100\%) $p=1.0$. Constant 0.7 and 0.5 schedules for Theseus, as well as linear 0.1--1.0 over 50\% of steps was attempted for Theseus as suggested by their paper and extrapolated to our training setup, but aggr20 was found to outperform for our experiments. DFG showed best results with matched schedule to aggr20.
\paragraph{Additional Details.}  
Key settings and terms used in experiments:  
\begin{itemize}
    \item \textbf{Gumbel}: Theseus-Gumbel stochastic replacement with temperature $\tau=1.0$.  
    \item \textbf{KL Distillation}: Teacher-student soft-target guidance with fixed temperature 4.  
    \item \textbf{DFG (Deep Feature Guidance)}: Auxiliary L2 loss on student-teacher intermediate outputs, controlled by weight $\lambda$.  
    \item \textbf{Student Initialization}: Reinitialized attention modules using Kaiming initialization.  
    \item \textbf{Hyperparameter Search}: Optimal settings determined via preliminary student-only cross-entropy training.  
\end{itemize}

\subsection{Theoretical Results and Proofs}
\label{app:theory}

This section contains the full statements and proofs of all theoretical propositions referenced in the main text.

\subsubsection{Proposition 1: Variance Decomposition for Theseus}
\label{app:proof-prop1}

\noindent\textbf{Proposition 1 (Variance decomposition for Theseus).}
If a hard gate $z\sim\mathrm{Bernoulli}(p)$ selects student vs.\ teacher (independent of the data), then
\[
\nabla_{\theta_\ell} L \;=\; z\,a,\qquad
\mathbb{E}[\nabla_{\theta_\ell}L] = p\,\mathbb{E}[a],
\]
\[
\mathrm{Var}[\nabla_{\theta_\ell}L] \;=\; p\,\mathrm{Var}[a] \;+\; p(1-p)\,\|\mathbb{E}[a]\|^2
\;\le\; p\,\mathrm{Var}[a] + p(1-p)\,\mathbb{E}\|a\|^2,
\]
where $a := J_{G_\ell}^\top \frac{\partial S_\ell}{\partial \theta_\ell} \in \mathbb{R}^{\dim(\theta_\ell)}$.

\noindent\emph{Proof.}
By independence, $\mathbb{E}[z a]=\mathbb{E}[z]\mathbb{E}[a]=p\,\mathbb{E}[a]$.
For the variance, by the law of total variance,
\begin{align*}
    \mathrm{Var}[z a] & \;=\; \mathbb{E}[\mathrm{Var}(z a \mid z)] + \mathrm{Var}(\mathbb{E}[z a \mid z]) \\
    & = \mathbb{E}[z^2]\,\mathrm{Var}[a] + \mathrm{Var}(z\,\mathbb{E}[a]) \\
    & = p\,\mathrm{Var}[a] + p(1-p)\,\|\mathbb{E}[a]\|^2.
\end{align*}
Use $\|\mathbb{E}[a]\|^2\le \mathbb{E}\|a\|^2$ for the inequality. \hfill$\square$

\subsubsection{Proposition 2: Deterministic Gate Removes Gate-Induced Variance}
\label{app:proof-prop2}

\noindent\textbf{Proposition 2 (Deterministic gate removes gate-induced variance).}
Let $Y_\alpha := (1-\alpha)\,T_\ell + \alpha\,S_\ell$ and define $a(y) := J_{G_\ell}(y)^\top \frac{\partial S_\ell}{\partial \theta_\ell}$, so that the DCR gradient is
\[
\nabla_{\theta_\ell} L_{\mathrm{DCR}} \;=\; \alpha\,a(Y_\alpha),
\]
whereas under Theseus (hard gate $z\!\sim\!\mathrm{Bernoulli}(p)$) the student gradient is $\,\nabla_{\theta_\ell} L_{\mathrm{Th}} = z\,a(S_\ell)$. Let $X$ denote the minibatch. Then the \emph{gate-induced} component of gradient variance,
\[
\mathbb{E}\!\left[\mathrm{Var}\!\big(\nabla_{\theta_\ell} L \,\big|\, X\big)\right],
\]
is zero for DCR and equals $p(1-p)\,\mathbb{E}\|a(S_\ell;X)\|^2$ for Theseus. Hence,
\[
\mathbb{E}\!\left[\mathrm{Var}\!\big(\nabla_{\theta_\ell} L_{\mathrm{Th}} \,\big|\, X\big)\right]
\;-\;
\mathbb{E}\!\left[\mathrm{Var}\!\big(\nabla_{\theta_\ell} L_{\mathrm{DCR}} \,\big|\, X\big)\right]
\;=\; p(1-p)\,\mathbb{E}\|a(S_\ell;X)\|^2 \;\ge\; 0.
\]

\noindent\emph{Proof.}
Condition on $X$. Under Theseus, $\nabla_{\theta_\ell} L_{\mathrm{Th}} = z\,a(S_\ell;X)$ with $z\!\perp\!X$, so $\mathrm{Var}(\nabla_{\theta_\ell} L_{\mathrm{Th}}\mid X)=p(1-p)\|a(S_\ell;X)\|^2$ and taking expectation over $X$ gives the stated value. Under DCR there is no gate randomness given $X$, so $\mathrm{Var}(\nabla_{\theta_\ell} L_{\mathrm{DCR}}\mid X)=0$. \hfill$\square$

\subsubsection{Proposition 3: Curvature Bias Bound}
\label{app:prop3}

\noindent\textbf{Proposition 3 (Curvature bias bound).}
Let $\psi:\mathbb{R}^d\to\mathbb{R}$ be twice differentiable with $\sup_{y\in \mathrm{seg}(T_\ell,S_\ell)} \|\nabla^2 \psi(y)\|_{\mathrm{op}} \le M$ (segment between $T_\ell$ and $S_\ell$), and set $\mu=(1-p)\,T_\ell+p\,S_\ell$, $\Delta=S_\ell-T_\ell$. For Theseus with $Y = z\,S_\ell + (1-z)\,T_\ell$ where $z\sim \mathrm{Bernoulli}(p)$, we have
\[
\big|\ \mathbb{E}[\psi(Y)] - \psi(\mu)\ \big|
\;\le\; \tfrac{M}{2}\, p(1-p)\, \|\Delta\|^2.
\]

\noindent\emph{Proof.}
Second-order Taylor around $\mu$ gives
$\psi(Y) = \psi(\mu) + \nabla \psi(\mu)^\top (Y-\mu)
+ \tfrac{1}{2}\,(Y-\mu)^\top \nabla^2 \psi(\xi_Y)\,(Y-\mu)$
for some $\xi_Y$ on the segment between $Y$ and $\mu$.
Take expectations: the linear term vanishes ($\mathbb{E}[Y-\mu]=0$), and
$\mathbb{E}\|Y-\mu\|^2=\mathbb{E}[(z-p)^2]\|\Delta\|^2=p(1-p)\|\Delta\|^2$. \hfill$\square$

\noindent\textbf{Corollary (Deterministic path avoids mixing bias).}
For DCR, $Y_\alpha=(1-\alpha)\,T_\ell+\alpha\,S_\ell$ is deterministic, so $\mathbb{E}[\psi(Y_\alpha)]=\psi(Y_\alpha)=\psi(\mathbb{E}[Y_\alpha])$; no stochastic mixing bias arises. Theseus pays a curvature-dependent penalty scaling with $p(1-p)\|\Delta\|^2$.

\subsubsection{Proposition 4: Smooth Loss Path}
\label{app:prop4}

\noindent\textbf{Proposition 4 (Smooth loss path).}
Let $f(y):=(L\!\circ\!G_\ell)(y)$. If $f$ is $L_y$-Lipschitz on the segment between $T_\ell$ and $S_\ell$, then along $y(\alpha)=(1-\alpha)\,T_\ell+\alpha\,S_\ell$,
\[
\big|\, f(y(\alpha)) - f(y(0)) \,\big|
\;\le\; L_y\, \alpha\, \|S_\ell-T_\ell\|
\;\le\; L_y\, \alpha\, D_\ell
\quad\text{for any } D_\ell \ge \|S_\ell-T_\ell\|.
\]

\noindent\emph{Proof.}
$y(\alpha)-y(0)=\alpha(S_\ell-T_\ell)$ and Lipschitz continuity give the bound. \hfill$\square$

\subsection{DCR Implementation Details}

\begin{algorithm}[h]
\caption{DCR training step with global gate $\alpha(t)$ (pre-norm residual).}
\label{alg:dcr_process}
\begin{algorithmic}[1]
\Require batch $(x,y^\star)$; frozen teachers $\{T_\ell\}_{\ell\in\mathcal{I}}$ (\texttt{eval()}, no-grad); trainable students $\{S_\ell\}_{\ell\in\mathcal{I}}$; global gate $\alpha(t)$
\State $x_1 \leftarrow x$
\For{$\ell=1$ \textbf{to} $L$}
  \If{$\ell\notin\mathcal{I}$} \State $x_{\ell+1}\!\leftarrow\!\text{original block forward}$
  \Else
    \State $h_\ell \leftarrow \mathrm{LN}(x_\ell)$
    \State \textbf{without gradients:} $t_\ell \leftarrow T_\ell(h_\ell)$
    \State $s_\ell \leftarrow S_\ell(h_\ell)$
    \State $x_{\ell+1} \leftarrow x_\ell + \alpha(t)\,t_\ell + (1-\alpha(t))\,s_\ell$
  \EndIf
\EndFor
\State $\hat{y} \leftarrow$ task head on $x_{L+1}$
\State $\mathcal{L} \leftarrow \mathcal{L}_{\mathrm{task}}(\hat{y},y^\star)$
\State Backprop (student params only); optimizer step; skip computing $t_\ell$ once $\alpha(t)= 0$
\end{algorithmic}
\end{algorithm}

\end{document}